\documentclass{article}

\newif\ifarxiv
\arxivtrue
\ifarxiv\PassOptionsToPackage{preprint}{neurips_2026}\fi
\usepackage[sglblindworkshop]{neurips_2026}
\usepackage[utf8]{inputenc}
\usepackage[T1]{fontenc}
\usepackage[hidelinks]{hyperref}
\usepackage{url}
\usepackage{booktabs}
\usepackage{amsfonts}
\usepackage{amsmath}
\usepackage{nicefrac}
\usepackage{microtype}
\usepackage{graphicx}
\usepackage{xcolor}

\workshoptitle{E-Values: From Statistics to ML}

\title{When the Martingale Never Stops Firing:\\Anytime-Valid Gating on Real Forecast Streams}

\author{%
  Weijia Han\\
  University of Washington\\
  Seattle, USA\\
  \texttt{wh73@uw.edu}\\
  \And
  Lisha Qu\\
  University of Washington\\
  Seattle, USA\\
  \texttt{lishaq@uw.edu}\\
}

\begin{document}

\maketitle

\begin{abstract}
Machine learning systems are increasingly corrected while they run, and the decision of
when to intervene is increasingly delegated to statistical monitors. Anytime-valid
inference promises evidence that can be acted on at any moment, exactly the guarantee
this setting needs, and it is moving from theory into deployed monitoring. Conformal test
martingales are the change-detection instrument, and Ville's inequality caps their
false-alarm probability on exchangeable data. The guarantee is conditional. A deployment inherits it only if the stream it monitors behaves
exchangeably. The premise is hardest to satisfy where these monitors are most useful, on
dependent data and inside loops where the monitor modifies the learner whose scores it
reads. It is also rarely measured. We measure it in a pre-specified case study, where
such a monitor gates the online updates of a Kalman adapter correcting frozen
time-series foundation models on five forecasting streams. On exchangeable synthetic
streams, the same implementation fires in at most \textbf{1 of 60} runs. On the real
streams, at $\alpha=0.05$, \textbf{135 of 135} clean-stream runs fired. The construction
does not explain the firing; the failure comes from the deployed score stream itself. Repeated fires hold
the gate's drift response active, and the gated filter amplifies the very transient it
was designed to prevent. The component worth keeping
makes no validity claim. Huber-style gating of the filter's own updates cuts
isolated-spike degradation by an order of magnitude with no dataset-specific tuning. Anytime-valid
methods proposed for dependent data should therefore be accompanied by
null-calibration controls and mechanism traces.
\end{abstract}

\section{A pre-specified martingale-gated deployment}
\label{sec:setting}

Anytime-valid monitors are advocated precisely for online deployment. They fire as soon as
evidence accumulates, at a finite-sample false-alarm level that holds at every stopping
time~\citep{ramdas2023game}. That level rests on a premise, exchangeability of the monitored
score stream. We measure that premise on a single deployment, frozen before any run.

\textbf{Backbones and adapter.} Frozen time-series foundation models, hereafter backbones, are
increasingly corrected
\emph{online} by lightweight adapters. We use TiRex~\citep{auer2025tirex},
Chronos-2~\citep{ansari2024chronos,ansari2025chronos2}, and
TimesFM~2.5~\citep{das2024decoderonly}, the last a checkpoint release of the cited model
line.
We correct each backbone with a per-step linear ``black-box stacking'' adapter (forecast horizon 96, state dimension 192)
maintained by a Kalman filter whose hyperparameters $(\lambda, Q, R)$ adapt online by Adam on
the innovation likelihood~\citep{westharrison1997}. An exact block-structured form makes
the filter and its hyper-gradient cost $0.163\times$ one-step SGD; the gate's conformal and
martingale path is not counted.

\begin{figure}[t]
\centering
\includegraphics[width=\linewidth]{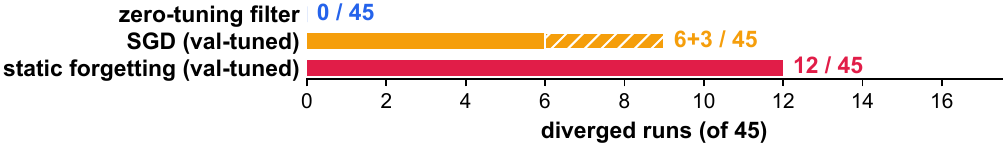}
\caption{Clean-stream divergences per learner across the 45 staged-study runs (5 datasets
$\times$ 3 backbones $\times$ 3 seeds). Hatching marks de facto divergences the magnitude
guard missed (SGD on traffic under TimesFM~2.5, cumulative MSE 7.97 against the committed
threshold of 5, no guard flag, in all three seeds).}
\label{fig:diverge}
\vspace{-5pt}
\end{figure}

\textbf{Motivation.} The staged study preceding the gate left exactly one
problem for a gate to solve. Validation-tuned
static forgetting diverges on test; its chosen cell finishes 1.4 to 3.6 orders of magnitude
worse than the zero-tuning filter on one dataset across all three backbones. The zero-tuning
adaptive filter never diverged in any of the 45 staged-study runs
(Figure~\ref{fig:diverge}). Its one
weakness is a single self-excited transient on the weather stream. Cumulative MSE stays
near $2\times10^{-4}$ through roughly step 18{,}400, a brief elevated-error episode
follows, and one event at step 18{,}536 dominates the final cumulative MSE.

This motivates a gate that (i) robustifies isolated spikes and (ii) declares sustained
drift so adaptation can be boosted, with a conformal test
martingale~\citep{vovk2003testing,vovk2021retrain,ville1939etude,shafer2021testing,ramdas2023game}
making the decision.

\textbf{Gate construction.} The nonconformity score is the filter's own standardized
innovation, and the monitor is a simple-mixture conformal test martingale over randomized
conformal $p$-values from a 250-step sliding calibration
window~\citep{vovk2022algorithmic}:
{\abovedisplayskip=4pt \belowdisplayskip=4pt \abovedisplayshortskip=2pt \belowdisplayshortskip=2pt
\begin{equation*}
s_t = e_t^{\top} S_t^{-1} e_t,
\qquad\quad
M_t = \frac{1}{K}\sum_{k=1}^{K}\;\prod_{\tau\le t} \varepsilon_k\, p_\tau^{\,\varepsilon_k-1},
\end{equation*}}
with $K{=}19$ grid values $\varepsilon_k\in\{0.05,\dots,0.95\}$, threshold
$\nicefrac{1}{\alpha}$ at $\alpha=0.05$, and reset on fire. A decision table maps each step to normal,
huberize~\citep{huber1964robust}, skip, or sustained drift: drift is declared on
$M_t\ge\nicefrac{1}{\alpha}$ (checked first), isolated spikes on $s_t$ above the $99.9$th
percentile of its fixed chi-square null distribution; the drift response boosts $Q$ by
$10\times$, decaying linearly over 50 recovery steps, applies a covariance soft reset, and
moves the forgetting factor to a drift value. Spike classification is suppressed during those
recovery steps.

\textbf{Protocol.} Five datasets~\citep{zhou2021informer,wu2021autoformer} (ETTh2, ETTm1,
weather, electricity, traffic; test blocks of 12{,}922 to 52{,}117 steps), three backbones,
three seeds, deterministic cached-forecast replay. The online protocol is the
immediate-reveal streaming convention: stride one, each step's full 96-step label revealed
before the next update, so consecutive labels overlap in 95 of 96 entries, which is itself a
mechanical source of serial dependence in the scores; the full-label reveal is optimistic
for real deployment, and clean means uncontaminated replay with no stationarity
claim. Label contamination is fraction-pinned
(exactly $\operatorname{round}(fT)$ steps at $f\in\{1,5,10\%\}$; spike, level-shift, and stuck
injectors), applied to the update channel only; accuracy is scored against clean labels. A
run that crosses a magnitude-divergence guard is truncated at that step and marked diverged.
All verdicts, exclusion rules, and contamination severities were version-stamped
before the runs; Appendix~\ref{app:repro} states the
composite criterion.

\section{Premise failure on real streams and its attribution}
\label{sec:premise}

\begin{table}[t]
\caption{Fires on \textbf{clean} streams at $\alpha=0.05$; each range spans 3 backbones
$\times$ 3 seeds (9 runs). Right column: largest Ljung-Box $p$ across the cell's full-gate
runs; the weather value is a lower-tail degeneracy (see text).
$^{\dagger}$Counts cover a shorter stream, truncated by clean-stream divergence (6 of 9
weather full-gate and 3 of 9 traffic SGD-surrogate runs, Section~\ref{sec:mechanism}).}
\label{tab:fires}
\vspace{-4pt}
\centering
\footnotesize
\begin{tabular}{lrrrr}
\toprule
dataset & Huber-only & full gate & SGD surrogate & max LB $p$ (full gate)\\
\midrule
ETTh2 & 31--72 & 18--35 & 121--173 & $<10^{-8}$\\
ETTm1 & 459--557 & 146--176 & 634--712 & $<10^{-15}$\\
weather & 462--567 & 115--277$^{\dagger}$ & 787--1{,}340 & $\approx 1$ (degenerate)\\
electricity & 77--94 & 36--56 & 57--75 & $0.25$\\
traffic & 26--43 & 22--33 & 118--210$^{\dagger}$ & $<10^{-10}$\\
\bottomrule
\end{tabular}
\end{table}

Three gated arms isolate the gate's components. The full gate is the deployed object.
Huber-only keeps the huberize and skip branches with the drift response disabled. The SGD surrogate drives the same gate from a
trailing mean-square score on an SGD learner.
Table~\ref{tab:fires} reports the primary count: across 45 clean-stream reports with 3 seeds each, \textbf{135 of 135 runs fired at least once}, all 45
full-gate runs among them (first fires at steps 6 to 4{,}274, median 460; 18 to 1{,}340 total fires). Under the premise and the textbook guarantee, Ville's bound
at $\alpha=0.05$ would cap the expected number of runs with any fire near 7 of 135. Every SGD-surrogate run fired, so firing is not specific
to the filter's scores.
Two explanations compete: the score
streams themselves, and the deployed deviations from the textbook protocol.

\textbf{Inconclusive whiteness diagnostics.}
Ljung-Box~\citep{ljungbox1978} at 10 lags, on the first-step ($h{=}1$) innovation column,
non-overlapping in target time, rejects at
$p<0.01$ in \textbf{119 of 135} runs, consistent with serially dependent innovations. But
the portmanteau over-rejects under conditional heteroskedasticity even for uncorrelated
series, which a shipped illustration demonstrates. The weather
full-gate runs report a portmanteau statistic between $0.05$ and $0.64$
against a $\chi^2_{10}$ null with mean 10, so $p\approx1$ by \emph{lower-tail} degeneracy, an
artifact of the very transient traced in Section~\ref{sec:mechanism}.

\textbf{Null controls of the construction.} The deployed construction
departs from the classical online protocol in three ways: a \emph{sliding} calibration window
in place of the growing bag, scores from an adaptively learned $S_t$, and reset on fire. We ran the null control: the same window, randomized $p$-values,
mixture, and reset logic on exchangeable i.i.d.\ score streams, under both a pre-filled
window and the deployed empty-window initialization, at $T{=}20{,}000$ and the longest deployed
$T{=}52{,}117$. The pre-filled conditions fired in \textbf{0 of 60} runs
each; the deployed initialization fired in 1 of 60 at either horizon; heavy-tailed
($|t_2|$) scores fired in 0 of 60. Sixty runs cannot certify strict conservatism. The observed fire fraction stayed below $\alpha$ in every condition, and reset on fire cannot affect the \emph{first} crossing, the event that Ville's
inequality bounds. On exchangeable inputs the implementation stays quiet, so the pervasive firing does
not come from the construction alone; the deployed variant carries no proven Ville
guarantee of its own.

A complementary control keeps the full deployed
pipeline, drift response included, and feeds it drift-free synthetic streams whose labels
are a smooth deterministic pattern plus i.i.d.\ Gaussian noise, so every monitored score
comes from the pipeline's own adaptation. Were the premise to hold on those scores, at
most 3 of 60 runs would be expected to fire. \textbf{60 of 60} fired, with 33 to 42 fires
per run and first fires inside 67 steps of the test block; the i.i.d.-score control shares
the deployed empty-window start and fired in 1 of 60, so the early fires are not a window
artifact. With no drift anywhere in the data, the pipeline's own scores break the
premise, though the control does not isolate the responsible element of the loop.

\section{Fire-triggered amplification of the weather transient}
\label{sec:mechanism}

\begin{figure}[t]
\centering
\includegraphics[width=\linewidth]{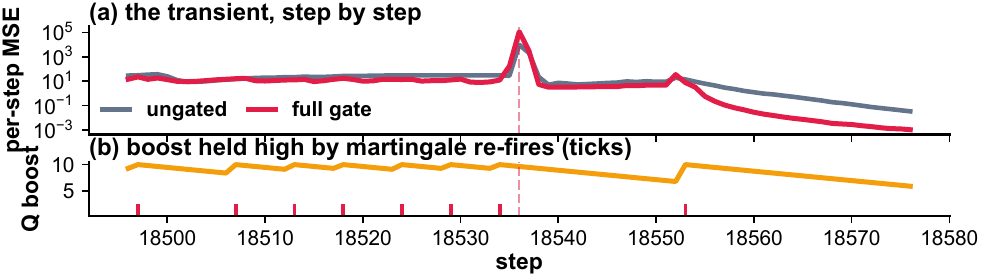}
\caption{Single-step trace around the natural weather transient (clean stream, TimesFM~2.5,
seed 0). \textbf{(a)} Per-step MSE: the full gate (red) tracks the ungated filter until the
event, then exceeds it by $12\times$. \textbf{(b)} $Q$-boost multiplier with fire ticks
(8 plotted of 264): re-fires hold an 8 to $10\times$ boost active at the event.}
\label{fig:transient}
\end{figure}

A premise failure matters only through what the fires trigger; we trace the worst case, the
self-excited weather transient of Section~\ref{sec:setting}.
Figure~\ref{fig:transient} shows that event at single-step resolution on TimesFM~2.5.
Through the elevated-error episode preceding the event, ungated per-step MSE runs 8 to 37
where it typically remains below 1. The martingale re-fires every 5 to 10 steps through that
episode, so the drift response's $Q$-boost never decays below 8. At the event step the configured boost
multiplier reads 9.6 and logged process noise is $2.0\times$ the ungated filter's.

Under the accumulated drift response, TimesFM~2.5's
per-step MSE at the event reaches $1.155\times10^{5}$ against the ungated filter's
$9.8\times10^{3}$, and its final cumulative MSE rises from 0.36 to between 2.6 and 3.1
across seeds. On TiRex and Chronos-2, where
amplification is larger, the same event crosses the magnitude-divergence guard \emph{on the
clean stream itself}, in six of six runs, all at step 18{,}536. It crosses again in all
\textbf{54 of 54} contaminated weather runs on those two backbones. Component and feedback controls in Appendix~\ref{app:ablations} attribute the divergence
to the covariance soft reset and to the immediate reveal; the boost alone never diverges.

The spike branch of the decision table never activated.
The recorded action at the event was a normal update, recovery-window suppression was
active, and the event's conformal $p$ of $0.011$ reflected a window already adapted to the
episode. The trace also exposed a configuration incoherence in the deployed gate. The base
forgetting factor
is pinned at the guard floor ($0.90$), \emph{below} the drift value ($0.95$), so the
drift response simultaneously slowed forgetting and boosted process noise.

\section{Component ablations under contamination and drift}
\label{sec:survives}

\begin{table}[t]
\caption{Isolated-spike label contamination ($5\%$ of steps, $6$ robust-$\sigma$): mean
per-run degradation $(\text{contaminated}-\text{clean})/\text{clean}$, paired within
(backbone, seed). Ungated arms are the same learners with no gate; cells losing runs to divergence show
$[n\ \text{used}/9]$.}
\label{tab:spike}
\centering
\footnotesize
\renewcommand{\arraystretch}{0.92}
\begin{tabular}{lrrrrr}
\toprule
dataset & ungated KF & Huber-only & full gate & ungated SGD & SGD surrogate\\
\midrule
ETTh2 & $+3.32$ & $+0.15$ & $+0.24$ & $+9.04$ & $+0.15$\\
traffic & $+3.73$ & $+0.11$ & $+0.35$ & $+2.07\,[3/9]$ & $+3.40\,[6/9]$\\
ETTm1 & $+2.68$ & $+0.63$ & $+6.28$ & $+2.90$ & $+0.06$\\
electricity & $+0.04$ & $+0.01$ & $-0.31$ & $+1.95$ & $+2.08$\\
weather & $+0.01$ & $-0.03\,[8/9]$ & $+0.03\,[3/9]$ & $+0.00$ & $-0.02$\\
\bottomrule
\end{tabular}
\vspace{-8pt}
\end{table}

The drift response caused the damage. The remaining question is which component is worth
keeping.
Table~\ref{tab:spike} isolates the robust-update half (huberize and skip). It delivers the
spike robustness without the drift response and matches or beats the full gate on ETTh2,
traffic, and ETTm1, the three datasets where the ungated learner degrades appreciably.
The same gating transfers to SGD through the surrogate score on ETTh2 and ETTm1. On traffic the
surrogate arm does worse than the ungated SGD learner. Divergence is itself an outcome, so
the bracketed survivor-only cells favor the arms that lose runs; we read them qualitatively. The Huber-only arm still uses the martingale, though only as a classifier, and nothing here needs an anytime-validity claim.

The same protocol also produced negative results. Sustained contamination defeats innovation
gating. Level shift at $5\%$ degrades every arm by $+7.8$ to $+19.3$ on ETTh2 and
ETTm1, with the gated arms often worst. On traffic it degrades every Kalman arm by $+9.5$
to $+18.3$. The
stuck injector degrades the full gate by $+1.7$ to $+14.2$ on the four datasets outside
weather, against at worst $+1.8$ for the ungated filter and $+0.5$ for the SGD arms. On
synthetic piecewise drift the full gate adapts \emph{more slowly} than the
ungated filter on three of three seeds (Appendix~\ref{app:latency}). The pre-specified
composite gate criterion (Appendix~\ref{app:repro}) fails on three of three backbones.

\section{Implications for e-value methods}
\label{sec:implications}

Deployment reports should pair every anytime-valid claim with raw fire counts on clean data, a
\emph{null-calibration control of the deployed implementation} at its horizon, and
dependence diagnostics. Guarantee-free components should be evaluated separately from
guaranteed ones. Here the robust-update half carried the measured benefit and the
martingale-triggered \emph{response} caused the damage, so an ``anytime-valid gating''
framing would credit the benefit to the guarantee and hide the source of the damage.

On the detection half, e-detectors offer sequential change detection under
average-run-length control~\citep{shin2024edetectors}, though
replayed on the deployed $p$-streams they alarm in 135 of 135 runs at every tested
threshold (Appendix~\ref{app:ablations}), so a detector swap does not repair the premise.
E-processes can test the exchangeability assumption itself~\citep{ramdas2022exchangeability}
and extend across filtrations~\citep{choe2024combining}, and adaptive conformal updating
tracks calibration under shift~\citep{gibbs2021adaptive}. None of these supplies a response policy with
bounded cost under wrong alarms, and that policy is what this study finds missing.

\bibliographystyle{plainnat}
\bibliography{refs}

\appendix

\section{Reproducibility and scope}
\label{app:repro}

We release deterministic cached-forecast replay, pre-specified verdicts with an
errata trail, raw counts beside every $p$-value, and the null-control scripts. All results
use the single gate configuration of Section~\ref{sec:setting}, so every claim here is a claim
about that deployment. Seeds are Monte Carlo replicates conditional on each fixed cached stream. Each gated arm draws its own
randomized conformal $p$-values per seed, and each contaminated run spawns its own RNG. The
pre-specified composite criterion requires the gated arm to beat the ungated arm, per
contamination kind, on at least three of five comparable dataset cells (cells where the
ungated arm itself degrades by less than five percent are excluded as incomparable); the
gate passes only if every evaluated kind passes. Contamination magnitudes are in units of a
robust per-horizon scale of the clean labels (1.4826 times the median absolute deviation).
Spike alters isolated uniformly chosen steps by six scale units with per-step random sign;
level-shift applies a three-unit offset over one contiguous block with one sign per
horizon; stuck freezes one contiguous block at its onset value with no magnitude at all. The TimesFM~2.5 backbone
is the \texttt{google/timesfm-2.5-200m-pytorch} checkpoint
release, pinned by revision alongside the other two backbones in the code release. Code and
artifacts will be released, anonymized during review.

\section{Component and feedback ablations}
\label{app:ablations}

All numbers replay the committed run artifacts; MSE figures quote TimesFM~2.5, the backbone
of the Section~\ref{sec:mechanism} trace, and divergence counts cover all three backbones.
\textbf{Response components.} On clean weather with immediate feedback the soft-reset-only
arm reproduces the full gate's divergence in six of six TiRex and Chronos-2 runs at step
18{,}536 and ends inside the full gate's range on TimesFM~2.5 (2.50 to 3.03 against 2.57 to
3.06), while the boost-only arm diverges in zero of nine runs and ends near the ungated
filter (0.378 to 0.383 against 0.361); the forgetting-only arm is likewise benign (0.341 to
0.342). The covariance soft reset is sufficient for the catastrophe; the boost alone is not.

\textbf{Delayed feedback.} With labels revealed 96 steps late the clean-weather full gate
diverges in zero of nine runs, and final cumulative MSE stays above the ungated filter on
every backbone (TiRex 0.50 to 0.53 against 0.286, Chronos-2 8.3 to 9.1 against 5.75,
TimesFM~2.5 1.18 to 1.23 against 0.517). The step-18{,}536 catastrophe requires the
immediate full-vector reveal.

\textbf{A martingale-free cutoff.} A fixed-threshold huberize-and-skip arm with frozen
hyperparameters and no martingale holds isolated-spike degradation to at most $+0.05$ in
every dataset, stronger than Huber-only on the three materially affected datasets. Its
clean-weather cost is backbone-dependent, from well below the ungated filter on TiRex and
Chronos-2 (0.37 against 2.03; 0.23 against 3.58) to $9.5\times$ above it on TimesFM~2.5
(3.41 against 0.36).

\textbf{E-detector replay.} SR and CUSUM e-detectors~\citep{shin2024edetectors} built from
the deployed betting family, replayed on the committed $p$-value streams, first reproduce
the recorded fires exactly in 405 of 405 runs. Both then alarm in 135 of 135 paper runs at
every threshold $A\in\{10^{3},10^{4},10^{5}\}$, with SR at $A{=}10^{4}$ producing
193/320/1{,}464 min/median/max alarms per run against null expected counts of 1.3 to 5.2,
while the same detectors stay calibrated on i.i.d.\ $p$-streams and on i.i.d.\ scores
passed through the deployed empty-start sliding window. The detector family inherits the
premise failure; the remedy must restore score exchangeability.

\section{Synthetic drift latency}
\label{app:latency}

Table~\ref{tab:latency} reports the synthetic-drift latency study over three seeds with
declared break points and a per-break bounded search. It gives the mean adaptation latency,
the number of steps after each declared break until a trailing-error recovery criterion is
met (never-recovered breaks are capped at the bounded search horizon), and the final
cumulative MSE per arm. The full gate is slower than the ungated filter at its own
motivating task on every seed, and its final error is roughly four times worse.

\begin{table}[h]
\centering
\caption{Mean latency (steps) and final cumulative MSE on the synthetic drift benchmark.}
\label{tab:latency}
\begin{tabular}{lrrrrrr}
\toprule
 & \multicolumn{3}{c}{mean latency} & \multicolumn{3}{c}{final MSE} \\
\cmidrule(lr){2-4} \cmidrule(lr){5-7}
arm & seed 0 & seed 1 & seed 2 & seed 0 & seed 1 & seed 2 \\
\midrule
ungated filter & 126.8 & 128.0 & 95.2 & 0.092 & 0.101 & 0.098 \\
full gate      & 154.8 & 137.5 & 128.0 & 0.342 & 0.447 & 0.365 \\
SGD ungated    & 226.0 & 250.0 & 216.8 & 0.327 & 0.333 & 0.330 \\
\bottomrule
\end{tabular}
\end{table}

\end{document}